\documentclass[11pt,a4paper]{article}

\usepackage[hyperref]{acl2020}
\usepackage{mathpartir}
\usepackage{hyperref}
\usepackage{comment}
\hypersetup{
    colorlinks=true,
    linkcolor=blue,
    filecolor=magenta,      
    urlcolor=blue,
    pdftitle={Overleaf Example},
    pdfpagemode=FullScreen,
    citecolor=blue,
    }
    
\usepackage{listings}
\usepackage{times}
\usepackage{latexsym}
\usepackage{amsmath}
\usepackage{graphicx}
\usepackage{float}
\usepackage{amsfonts}
\usepackage{amssymb}
\usepackage{booktabs, tabularx}
\usepackage{multicol}
\usepackage{siunitx}
\usepackage{nameref}

\usepackage{microtype}
\usepackage{makecell}
\usepackage{CJKutf8}
\usepackage{tabulary}
\usepackage{caption}

\hypersetup{linktocpage}
\hypersetup{
    colorlinks=true, 
    linktoc=all,     
}

\aclfinalcopy 

\newcolumntype{L}{>{\centering\arraybackslash}m{3cm}}

\usepackage{cleveref}
\begin{document}

\title{Evaluation of Semantic Answer Similarity Metrics}
 


\title{Evaluation of Semantic Answer Similarity Metrics} 
\author{\textsuperscript{1}Farida Mustafazade \\
  \textsuperscript{1}GAM Systematic\\
  \texttt{\small farida.mustafazade.15@ucl.ac.uk} \vspace{-2cm}
\And \textsuperscript{2}Peter Ebbinghaus \\
\textsuperscript{2}Teufel Audio\\
  \texttt{\small peter.ebbinghaus@posteo.de}\\\vspace{-2cm}
 }



\vspace{-2.5cm}
\maketitle
\begin{abstract} 
There are several issues with the existing general machine translation or natural language generation evaluation metrics, and  question-answering (QA) systems are indifferent in that context. To build robust QA systems,
we need the ability to have equivalently robust evaluation systems to verify whether model predictions to questions are similar to ground-truth annotations. The ability to compare similarity based on semantics as opposed to pure string overlap is important to compare models fairly and to indicate more realistic acceptance criteria in real-life applications. We build upon the first to our knowledge paper that uses transformer-based model metrics to assess semantic answer similarity and achieve higher correlations to human judgement in the case of no lexical overlap. 
We propose cross-encoder augmented bi-encoder and BERTScore models for semantic answer similarity, trained on a new dataset consisting of name pairs of US-American public figures. 
As far as we are concerned, we provide the first dataset of co-referent name string pairs along with their similarities, which can be used for training.

\end{abstract}

\section{Introduction}

\label{intro}




Having reliable metrics for evaluation of language models in general, and models solving difficult question answering (QA) problems, is crucial in this rapidly developing field. These metrics are not only useful to identify issues with the current models, but they also influence the development of a new generation of models. In addition, it is preferable to have an automatic, simple metric as opposed to expensive, manual annotation or a highly configurable and parameterisable metric so that the development and the hyperparameter tuning do not add more layers of complexity.
SAS, a cross-encoder-based metric for the estimation of semantic answer similarity \cite{risch2021semantic},
provides one such metric to compare answers based on semantic similarity. 

The central objective of this research project is to analyse pairs of answers similar to the one in \autoref{fig:ex_answer_improv_1} and to evaluate evaluation errors across datasets and evaluation metrics.
\begin{figure}  
\centering
\small
\fbox{
\begin{minipage}{23em}
\textbf{Question}: Who makes more money: NFL or Premier League? \\
\textbf{Ground-truth answer}: National Football League\\
\textbf{Predicted Answer}: the NFL  \\
\textbf{EM}: $0.00$  \\
$\mathbf{F_1}$: $0.00$  \\
\textbf{Top-1-Accuracy}: $0.00$\\  
\textbf{SAS}: $ 0.9008$  \\
\textbf{Human Judgment}: $2$ (definitely correct prediction)  \\
$\mathbf{f_{BERT}}$: $ 0.4317$\\	
$\mathbf{f^{\prime}_{BERT}}$: $0.4446$\\	
\textbf{Bi-Encoder}: 0.5019
\end{minipage}}
\caption{Representative example from NQ-open of a question and all semantic answer similarity measurement results.}
    \label{fig:ex_answer_improv_1}

\end{figure}

The main hypotheses that we will aim to test thoroughly through experiments are twofold. Firstly, lexical-based metrics are not well suited for automated QA model evaluation as they lack a notion of context and semantics.  
Secondly, most metrics, specifically SAS and BERTScore, as described in \cite{risch2021semantic}, find some data types more difficult to assess for similarity than others. 

After familiarising ourselves with the current state of research in the field in \autoref{related_work}, we describe the datasets provided in \cite{risch2021semantic} and the new dataset of names that we purposefully tailor to our model in \autoref{data}. 
This is followed by \autoref{models}, introducing the four new semantic answer similarity approaches described in \cite{risch2021semantic}, our fine-tuned model as well as three lexical n-gram-based automated metrics. Then in \autoref{analysis}, we thoroughly analyse the evaluation datasets described in the previous section and conduct an in-depth qualitative analysis of the errors. 
Finally, in \autoref{conclusion}, we summarise our contributions. 



\section{Related work}
\label{related_work}
We define semantic similarity as different descriptions for something that has the same meaning in a given context, following largely \cite{Zeng2007SemanticRB}'s definition of semantic and contextual synonyms.
%
In \citet{pmlr-v133-min21a}, 
the human annotators attach a label \texttt{2} to all predictions that 
are "definitely correct", \texttt{1} - "possibly correct", and \texttt{0} - "definitely incorrect".
Automatic evaluation based on exact match (EM) fails to capture semantic similarity for definitely correct answers, where 60\% of the predictions are semantically equivalent to
the ground-truth answer. Just under a third of the predictions that do not match the ground-truth labels were nonetheless correct. They also mention other reasons for failure to spot equivalence, such as time-dependence of the answers or underlying ambiguity in the questions. 

QA evaluation metrics in the context of SQuAD v1.0 \cite{rajpurkar2016squad} dataset are analysed in \citet{bulian2022tomayto}. They thoroughly discuss the limitations of EM and F1 score from n-gram based metrics, as well as the importance of context including the relevance of questions to the interpretation of answers. A BERT matching metric (Bert Match) is proposed for answer equivalence prediction, which performs better when the questions are included alongside the two answers, but appending contexts didn't improve results. Additionally, authors demonstrate better suitability of Bert Match in constructing top-\textit{k} model's predictions. In contrast, we will cover multilingual datasets, as well as more token-level equivalence measures, but limit our focus on similarity of answer pairs without accompanying questions or contexts.

Two out of four semantic textual similarity (STS) metrics that we analyse and the model that we eventually train depend on bi-encoder and BERTScore \cite{zhang2019bertscore}. The bi-encoder approach model is based on the Sentence Transformer structure \cite{reimers-gurevych-2019-sentence}, which is a faster adaptation of BERT for the semantic search and clustering type of problems. BERTScore uses BERT to generate contextual embeddings, then match the tokens of the ground-truth answer and prediction, 
followed by creating a score from the maximum cosine similarity of the matched tokens. This metric is not one-size-fits-all. On top of choosing a suitable contextual embedding and model, there is an optional feature of importance weighting using inverse document frequency (idf). The idea is to limit the influence of common words. One of the findings is that most automated evaluation metrics demonstrate significantly better results on datasets without adversarial examples, even when these are introduced within the training dataset, while the performance of BERTScore suffers only slightly.
\cite{zhang2019bertscore} uses machine translation (MT) and image captioning tasks in experiments and not QA.
\cite{chen-etal-2019-evaluating} apply BERT-based evaluation metrics for the first time in the context of QA. Even though they find that METEOR as an n-gram based evaluation metric proved to perform better than the BERT-based approaches, they encourage more research in the area of semantic text analysis for QA. Moreover, \cite{bulian2022tomayto} uses only BERTScore base as one of the benchmarks, while we explore the larger model, as well as a finetuned variation of it.

Authors in \cite{risch2021semantic} expand on this idea and further address the issues with existing general MT, natural language generation (NLG), which entails as well generative QA and extractive QA evaluation metrics. These include
reliance on string-based methods, such as EM, F1-score, and top-n-accuracy. 
The problem is even more substantial for multi-way annotations. Here, multiple ground-truth answers exist in the document for the same question, but only one of them is annotated. 
The major contribution of the authors is the formulation and analysis of four semantic answer similarity approaches that aim to resolve to a large extent the issues mentioned above. They also release two three-way annotated datasets: a subset of the English SQuAD dataset \cite{rajpurkar2018know}, German GermanQuAD dataset \cite{moeller2021germanquad}, and NQ-open \cite{pmlr-v133-min21a}.

As depicted in \autoref{categories} and \autoref{analysis}, the leading problematic data type category is entities, particularly those involving names. \citet{si2021s} analyse Natural Questions (NQ) \cite{47761}, TriviaQA \cite{DBLP:journals/corr/JoshiCWZ17} as well as SQuAD and address the issue that current QA benchmarks neglect the possibility of multiple correct answers. They focus on the variations of names, e.g. nicknames, and improve the evaluation of Open-domain QA models based on a higher EM score
by augmenting ground-truth answers with aliases from Wikipedia and Freebase. In our work, we focus solely on the evaluations of answer evaluation metrics and generate a standalone names dataset from another dataset, described in greater detail in \autoref{data}. 

Our main assumption is that better metrics will have a higher correlation with human judgement, but the choice of a correlation metric is important.
Pearson correlation is a commonly used metric in evaluating semantic text similarity (STS) for comparing the system output to human evaluation.  \cite{reimers-etal-2016-task} show that Pearson power-moment correlation can be misleading when it comes to intrinsic evaluation. They further go on to demonstrate that no single evaluation metric is well suited for all STS tasks, hence evaluation metrics should be chosen based on the specific task. In our case, most of the assumptions, such as  normality of data and continuity of the variables behind Pearson correlation do not hold. Kendall's rank correlations are meant to be more robust and slightly more efficient in comparison to Spearman as demonstrated in \cite{Croux2010spearmankendall}.





Soon after Transformers took over the field, adversarial tests resulted in significantly lower performance figures, which increased the importance of adversarial attacks \cite{niven-kao-2019-probing}.
General shortcomings of language models and their benchmarks led to new approaches such as Dynabench \cite{kiela2021dynabench} and AdvGLUE \cite{wang2021adversarial}.
There are other shortcomings of large language models, including environmental and financial costs \cite{10.1145/3442188.3445922}. 

\section{Data}
\label{data}
We perform our analysis on three \href{https://semantic-answer-similarity.s3.amazonaws.com/data.zip}{subsets} of larger datasets annotated by three human raters and provided by \cite{risch2021semantic}. 
Unless specified otherwise, these will be referred to by their associated dataset names. 


\subsection{Original datasets}

\textbf{SQuAD} is an English-language dataset containing multi-way annotated questions with 4.8 answers per question on average. \textbf{GermanQuAD} is a three-way annotated German-language question/answer pairs dataset created by the deepset team which also wrote \cite{risch2021semantic}. Based on the German counterpart of the English Wikipedia articles used in SQuAD, GermanQuAD is the SOTA dataset for German question answering models. To address a shortcoming of SQuAD that was mentioned in \cite{kwiatkowski-etal-2019-natural}, GermanQuAD was created with the goal of preventing strong lexical overlap between questions and answers. Hence, more complex questions were encouraged, and questions were  rephrased with synonyms and altered syntax.
SQuaD and GermanQuAD contain a pair of answers and a hand-labelled annotation of \texttt{0} if answers are completely dissimilar, \texttt{1} if answers have a somewhat similar meaning, and \texttt{2} if the two answers express the same meaning. 
\textbf{NQ-open} is a five-way annotated open-domain adaption of \citet{kwiatkowski-etal-2019-natural}'s Natural Questions dataset. NQ-open is
based on actual Google search engine queries.
In case of NQ-open, the labels follow a different methodology as described in \cite{pmlr-v133-min21a}.
The assumption is that we only leave questions with a non-vague interpretation (see \autoref{categories}). 
Questions like \textit{Who won the last FIFA World Cup?} received the label \texttt{1} because they have different correct answers without a precise answer at a point in time later than when the question was retrieved. There is yet another ambiguity with this question, which is whether it is discussing FIFA Women’s World Cup or FIFA Men’s World Cup. This way, the two answers can be correct without semantic similarity even though only one correct answer is expected.

The annotation of NQ-open indicates truthfulness of the predicted answer, whereas for SQuAD and GermanQuAD the annotation relates to the semantic similarity of both answers which can lead to differences in interpretation as well as evaluation.
To keep the methodology consistent and improve NQ-open subset, vague questions with more than one ground-truth labels have been filtered out. We also manually re-label incorrect labels as well as filter out vague questions. 


\autoref{dataset_stats} describes the size and some lexical features for each of the three datasets. 
There were 2, 3 and 23 duplicates in each dataset respectively. Dropping these duplicates led to slight changes in the metric scores.


\begin{table}[ht]
\centering 
\footnotesize
\setlength{\tabcolsep}{0.5\tabcolsep}

\begin{tabular}{@{}lrrr@{}}
\toprule
                    & \textbf{SQuAD} & \textbf{GermanQuAD} & \textbf{NQ-open} \\ \midrule
\textbf{Label 0}               & 56.7& 27.3& 71.7    \\
\textbf{Label 1}              & 30.7& 51.5& 16.6  \\
\textbf{Label 2 }                & 12.7& 21.1& 11.7    \\
$\mathbf{F_1 = 0}$                    & 565 & 124      & 3030 \\
$\mathbf{F_1 \neq 0}$                  & 374 & 299      & 529   \\
\textbf{Size}                & 939 & 423      & 3559 \\
\textbf{Avg answer size} & 23  & 68       & 13    \\ \bottomrule
\end{tabular}
\caption{Percentage distribution of the labels and statistics on the subsets of datasets used in the analyses. The average answer size column refers to the average of both the first and second answers as well as ground-truth answer and predicted answer (NQ-open only). $F_1=0$ indicates no string similarity, $F_1\neq0$ indicates some string similarity. Label distribution is given in percentages.}

\label{dataset_stats}

\end{table}
\subsection{Augmented dataset}
For NQ-open, the largest of the three datasets, names was the most challenging category to predict similarity as per \ref{fig:explanations}. While it includes city and country names as well, we focus on the names of public figures in our work. 
To resolve this issue, we provide a new dataset that consists of $\thicksim$40,000 (39,593) name pairs and employ the Augmented SBERT approach \cite{thakur2010augmentedsbert}: we use the cross-encoder model to label a new dataset consisting of name pairs and then train a bi-encoder model on the resulting dataset. We discuss the deployed models in more detail in \autoref{models}.
 
The underlying dataset is created from an \href{https://github.com/Kandy16/people-networks/tree/dbpedia-data/dbpedia-data/final_datasets}{open dbpedia-data dataset} \cite{wagner2017persondata} which includes the names of more than a million public figures that have a page on Wikipedia and DBpedia, including actors, politicians, scientists, sportsmen, and writers. Out of these we only use those with a U.S. nationality as 
the questions in NQ-open are on predominantly U.S. related topics.
We then shuffle the list of 25,462 names and pair them randomly to get the name pairs that are then labelled by the cross-encoder model.
It includes different ways of writing a person's name including aliases. 
For example, \textit{Gary A Labranche} and \textit{Labranche Gary}, or aliases like \textit{Lisa Marie Abato}'s stage name \textit{Holly Ryder} as well as e.g. Chinese ways of writing such as \textit{Rulan Chao Pian} and 
\begin{CJK*}{UTF8}{bsmi}
卞趙如蘭.
\end{CJK*}
We filter out all examples where more than three different ways of writing a person's name exist because in these cases these names don't refer to the same person but were mistakenly included in the dataset.
For example, names of various members of Tampa Bay Rays minor league who have one page for all members.
Since most public figures in the dataset have a maximum of one variation of their name, we only leave out close to 800 other variations this way, and can add 14,131 additional pairs.
These are labelled as 1 because they
refer to the same person.

\section{Models / Metrics}
\label{models}



The baseline semantic similarity models considered are bi-encoder, BERTScore vanilla, and BERTScore trained, whereas the focus will be on cross-encoder (SAS) performance. 
\autoref{configs} outlines the exact configurations used for each model. 
A cross-encoder architecture \cite{humeau2020polyencoders} concatenates two sentences with a special separator token and passes them to a network to apply multi-head attention over all input tokens in one pass. Pre-computation is not possible with the cross-encoder approach because it takes both input texts into account at the same time to calculate embeddings. A well-known language model that makes use of the cross-encoder architecture is BERT \cite{DBLP:journals/corr/abs-1810-04805}. The resulting improved performance in terms of more accurate similarity scores for text pairs comes with the cost of higher time complexity, i.e. lower speed, of cross-encoders in comparison to bi-encoders. A bi-encoder calculates the embeddings of the two input texts separately by mapping independently encoded sentences for comparison to a dense vector space which can then be compared using cosine similarity. The separate embeddings result in higher speed but reduced scoring
\cite{chen-etal-2020-dipair}. In our work, both cross- and bi-encoder architectures are based on Sentence Transformers \cite{DBLP:journals/corr/abs-1908-10084}.


\citet{risch2021semantic} used a separate English and German model for the cross-encoder because there is no multi-lingual cross-encoder implementation available yet. 
We use 
BERTScore implementation from \cite{zhang2019bertscore}
For BERTScore trained, the last layer representations were used, while for vanilla type BERTScore, only the second layer as per \autoref{embedding_extraction}.
BERTScore vanilla is based on 
\href{https://huggingface.co/bert-base-uncased}{bert-base-uncased} for English (SQuAD and NQ-open) and deepset's \href{https://huggingface.co/deepset/gelectra-base}{gelectra-base} \cite{chan-etal-2020-germans} for German (GermanQuAD), whereas BERTScore trained is based on the \textit{multi-lingual} model that is used by the bi-encoder \cite{may}.
\phantomsection
\label{new_model}
BERTScore trained outperforms SAS for answer-prediction pairs without lexical overlap, the largest group in NQ-open, but neither of the models perform well on names. 
New name pairs are used to train the Sentence Transformer, which can be found in \cite{sas2021evaluation}.



\section{Analysis}
\label{analysis}


To evaluate the shortcomings of lexical-based metrics in the context of QA, we compare BLEU, ROUGE-L, METEOR, $F_1$ and the semantic answer similarity metrics, i.e. Bi-Encoder, BERTScore vanilla, BERTScore trained, and Cross-Encoder (SAS) scores on evaluation datasets.
To address the second hypothesis, we delve deeply into every single dataset and analyse for disagreements with human judgement. 
As can be observed from \autoref{squad_and_nq_open} and \autoref{german_quad_correlations}, lexical-based metrics show considerably lower results than any of the semantic similarity approaches.

\begin{table}[ht]
\centering
\setlength{\tabcolsep}{.35\tabcolsep}
\footnotesize
\begin{tabular}{lllllllll}
\toprule
& \multicolumn{4}{c}{\textbf{SQuad}} & \multicolumn{4}{c}{\textbf{NQ-open}}\\
    \cmidrule(lr){2-5} \cmidrule(l){6-9}

& \multicolumn{2}{c}{\textbf{$F_1 = 0$}} & \multicolumn{2}{c}{\textbf{$F_1 \neq 0$}}& \multicolumn{2}{c}{\textbf{$F_1 = 0$}} & \multicolumn{2}{c}{\textbf{$F_1 \neq 0$}}\\
    \cmidrule(lr){2-3} \cmidrule(l){4-5}     \cmidrule(lr){6-7} \cmidrule(l){8-9}

        \textbf{Metrics}       & $\rho$& $\tau$& $\rho$& $\tau$   & $\rho$ & $\tau$ & $\rho$ & $\tau$  \\ \midrule
BLEU           & 0.00 & 0.00 & 0.17 & 0.16 & 0.00 & 0.00 & 0.05 & 0.05 \\
ROUGE-L        & 0.04 & 0.04 & 0.54 & 0.46 & 0.16 & 0.16 & 0.46 & 0.38 \\
METEOR         & 0.21 & 0.20 & 0.46 & 0.38 & 0.15 & 0.15 & 0.18 & 0.14 \\
F1-score       & 0.00 & 0.00 & 0.58 & 0.50 & 0.00 & 0.00 & 0.41 & 0.34 \\
Bi-Encoder     & 0.37 & 0.30 & 0.68 & 0.57 & 0.21 & 0.17 & 0.45 & 0.35 \\
$f_{BERT}$         & 0.13 & 0.11 & 0.60 & 0.49 & 0.17 & 0.14 & 0.14 & 0.11 \\
$f^{\prime}_{BERT}$          & 0.39 & 0.32 & 0.69 & 0.57 & 0.23 & 0.18 & 0.45 & 0.35 \\
SAS            & 0.36 & 0.29 & \textbf{0.74} & \textbf{0.61} & 0.20 & 0.16 & \textbf{0.65} & \textbf{0.51} \\
New Bi-Encoder & 0.39 & 0.32 & 0.69 & 0.57 & 0.25 & 0.20 & 0.50 & 0.39 \\
$\tilde{f}_{BERT}$          & \textbf{0.40} & \textbf{0.32} & 0.70 & 0.58 & \textbf{0.26} & \textbf{0.21} & 0.51 & 0.40 \\ \bottomrule
\end{tabular}
\caption{Spearman's, and Kendall's rank correlations of annotator labels and automated metrics on subsets of SQuAD and NQ-open. $f_{BERT}$ is BERTScore vanilla and $f^{\prime}_{BERT}$ is BERTScore trained, and $\tilde{f}_{BERT}$ is the new BERTScore trained on names.\\} 

\label{squad_and_nq_open}
\end{table}

BLEU lags behind all other metrics, followed by METEOR. Similarly, we found that ROUGE-L and F1 achieve close results.
In the absence of lexical overlap, METEOR gives superior results than the other n-gram-based metrics in the case of SQUAD, but ROUGE-L is closer to human judgement for the rest. 
The highest correlations are achieved in the case of BERTScore based trained models, followed closely by bi- and cross-encoder models. 
The superior performance of SAS doesn't hold up for the correlation metrics other than Pearson. We observed that SAS score underperformed when $F_1 = 0$ compared to all other semantic answer similarity metrics and overperformed when there is some lexical similarity. 

NQ-open is not only by far the largest of the three datasets but also the most skewed one. We observe that the vast majority of answer-prediction pairs have a label \texttt{0} (see \autoref{dataset_stats}). In the majority of cases, the underlying QA model predicted the wrong answer.

All four semantic similarity metrics perform considerably worse on NQ-open than on SQuAD and GermanQuAD. In particular, answer-prediction pairs that have no lexical overlap ($F_1=0$) amount to 95 per cent of all pairs with the label \texttt{0} indicating incorrect predictions. Additionally, they perform only marginally better than METEOR or ROUGE-L.


\begin{table}[ht]
\setlength{\tabcolsep}{.75\tabcolsep}
\footnotesize
\centering 
\begin{tabular}{@{}lllllll@{}}

\toprule
& \multicolumn{6}{c}{\textbf{GermanQuAD}}\\
& \multicolumn{3}{c}{\textbf{$F_1 = 0$}} & \multicolumn{3}{c}{\textbf{$F_1 \neq 0$}} \\
 \cmidrule(lr){2-4} \cmidrule(lr){5-7}
 \textbf{Metrics}           & $r$        & $\rho$& $\tau$& $r$        & $\rho$& $\tau$\\ \midrule
BLEU              & 0.000      & 0.000 & 0.000 & 0.153      & 0.095 & 0.089 \\
ROUGE-L           & 0.172      & 0.106 & 0.100 & 0.579      & 0.554 & 0.460 \\
$F_1$-score          & 0.000      & 0.000 & 0.000 & 0.560      & 0.534 & 0.443 \\
Bi-Encoder        & 0.392      & 0.337 & 0.273 & 0.596      & 0.595 & 0.491 \\
$f_{BERT}$  & 0.149      & 0.008 & 0.006 & 0.599      & 0.554 & 0.457 \\
$f^{\prime}_{BERT}$ & 0.410      & 0.349 & 0.284 & 0.606      & 0.592 & 0.489 \\
SAS                & \textbf{0.488}      & \textbf{0.432} & \textbf{0.349} & \textbf{0.713}      & \textbf{0.690} & \textbf{0.574} \\ \bottomrule
\end{tabular}
\caption{Pearson, Spearman's, and Kendall's rank correlations of annotator labels and automated metrics on subsets of GermanQuAD. $f_{BERT}$ is BERTScore vanilla and $f^{\prime}_{BERT}$ is BERTScore trained.}
\label{german_quad_correlations}

\end{table}

In \textbf{SQuAD}, there are only 16 cases where SAS completely diverges from human labels. In all seven cases where SAS score is above 0.5 and label is 0, we notice that the two answers have either \textbf{a common substring} or could be used often in the same context. In the other 9 extreme cases when the label is indicative of semantic similarity and SAS is giving scores below 0.25, there are three \textbf{spatial translations}.
There is an encoding-related example with 12 and 10 special characters each which seems to be a mislabelled example. 

Overall, error analysis for GermanQuAD is limited to a few cases because it is the smallest dataset of the three and all language model based metrics perform comparably well - SAS in particular. Regardless, SAS fails to identify semantic similarity in cases where the answers are \textbf{synonyms or translations} which also include technical terms that rely on Latin.
This is likely the case because SAS does not use a multilingual model.
Text-based \textbf{calculations and numbers} are also problematic.
SAS also fails to recognise \textbf{aliases or descriptions of relations} that point to the same person or object.

We also observe that similarity scores for answer-prediction pairs which include numbers, e.g. an amount, a date or a year, SAS, as well as BERTScore trained, 
diverge from labels.
The only semantically similar entities to answers expected to contain a numeric value should be the exact value, not a unit more or less. Also, the position within the pairs seems to matter for digits and their string representation. 
For SAS that the pair of \textit{11} and \textit{eleven} has a score of 0.09 whereas the pair of \textit{eleven} and \textit{11} has a score of 0.89. 

\begin{figure}
    \centering
    \includegraphics[width=.45\textwidth]{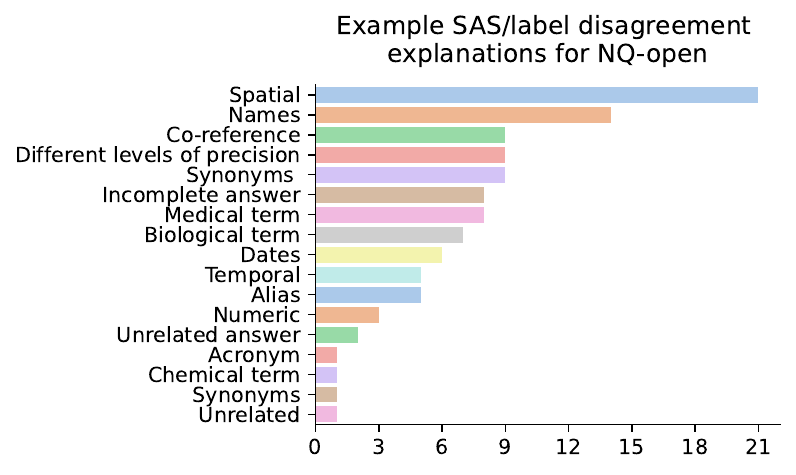}
    \caption{Subset of NQ-open test set, where SAS score $<$ 0.01 and human label is 2, manually annotated for an explanation of discrepancies. Original questions and Google search has been used to assess the correctness of the gold labels.}
    \label{fig:explanations}
\end{figure}

\autoref{fig:explanations} depicts the major error categories for when SAS scores range below 0.25 while human annotations indicate a label of 2. We observe that entities related to names, which includes spatial names as well as co-references and synonyms, form the largest group of scoring errors. 
After correcting for encoding errors and fixing the labels manually in the NQ-open subset, totalling 70 samples, the correlations have already improved by about a per cent for SAS. Correcting wrong labels in extreme cases where SAS score is below 0.25 and the label is \texttt{2} or when SAS is above 0.5 and label is \texttt{0} improves results almost across the board for all models, but more so for SAS.

After removal of duplicates, sample with imprecise questions, wrong gold label or multiple correct answers, we are left with 3559 ground-truth answer/prediction pairs compared to 3658 we started with.

An example for the better performance on names when applying our new bi-encoder and SBERT trained models can be seen in \autoref{fig:ex_answer_improv}, where both models perform well in comparison to SAS and human judgement.

\section{Conclusion}
\label{conclusion}

Existing evaluation metrics for QA models have various limitations. N-gram based metrics suffer from asymmetry, strictness, failure to capture multi-hop dependencies and penalise semantically-critical ordering, failure to account for relevant context or question, to name a few.
We have found patterns in the mistakes that SAS was making. These include \textbf{spatial awareness},  \textbf{names}, \textbf{numbers}, \textbf{dates},
\textbf{context awareness}, \textbf{translations}, \textbf{acronyms}, \textbf{scientific terminology}, \textbf{historical events}, \textbf{conversions}, \textbf{encodings}.

The comparison to annotator labels is performed on answer pairs taken from subsets of SQuAD and GermanQuAD datasets, and for NQ-open we have a prediction and ground-truth answer pair.
For cases with lexical overlap, ROUGE-L achieves comparative results to pre-trained semantic similarity evaluation models at a fraction of computation costs that the other models require. This holds for all GermanQuAD, SQuAD and NQ-open alike, discussed in more detail in \autoref{complexity}. 
Dataset size was one of the reasons why we focused more heavily on the NQ-open dataset. 
In addition, focusing on the other two would mean less strong evidence on how the metric will perform when applied to model predictions behind a real-world application. Furthermore, all semantic similarity metrics failed to have a high correlation to human labels when there was no token-level overlap, which is arguably the most important use-case for a semantic answer similarity metric as opposed to, say, ROUGE-L. NQ-open happened to have the largest number of samples that satisfied this requirement. 
Removing duplicates and re-labelling 
led to significant improvements across the board.
We have generated a names dataset, which was then used to fine-tune the bi-encoder and BERTScore model. The latter achieves and beats SOTA rank correlation figures when there is no lexical overlap
for datasets with English as the core language. 
Bi-encoders outperformed cross-encoders on answer-prediction pairs without lexical overlap both in terms of correlation to human judgement and speed, which makes them more applicable in real-world scenarios.

An element of future research would be further improving the performance on names of public figures as well as spatial names like cities and countries. 
Knowledge-bases, such as Freebase or Wikipedia, as explored in \cite{si2021s}, could be used to find an equivalent answer to named geographical entities. Numbers and dates which is the problematic data type in multi-lingual, as well as monolingual contexts, would be another dimension.  

\section*{Acknowledgements}

We would like to thank  Ardhendu Singh, Julian Risch, Malte Pietsch and XCS224U course facilitators, Ankit Chadha in particular, as well as Christopher Potts for their constant support.

\bibliography{references}
\bibliographystyle{acl_natbib}

\appendix
\pagebreak
\section{Error categories}

\begin{table*}[ht]
\centering
\setlength{\tabcolsep}{.3\tabcolsep}
\small
\begin{tabular}{@{}p{1.5cm}p{4.5cm}p{4cm}p{2cm}p{2cm}@{}}
\textbf{Category }                     & \textbf{Definition }                                                                                                                                           & \textbf{Question}                                                     & \textbf{Gold label}                    & \textbf{Prediction}                         \\ \toprule

Acronym                       & An abbreviation formed from the initial letters of other words and   pronounced as a word                                                            & what channel does the haves and have nots come on on directv & OWN                           & Oprah Winfrey Network              \\\midrule
Alias                         & Indicate an additional name that a person sometimes uses                                                                                              & who is the man in black the dark tower                       & Randall Flagg                 & Walter Padick                      \\\midrule
Co-reference                  & \begin{tabular}[c]{@{}l@{}}Requires resolution of a   relationship\\      between two distinct words referring\\      to the same entity\end{tabular} & who is marconi in we built this city                         & the father of the radio       & Italian inventor Guglielmo Marconi \\\midrule
Different levels of precision & When both answers are correct, but one is more precise                                                                                                & when does the sympathetic nervous system be activated        & constantly                    & fight-or-flight response           \\\midrule
Imprecise question            & There can be more than one correct answers                                                                                                            & b-25 bomber accidentally flew into the empire state building & Old John Feather Merchant     & 1945                               \\\midrule
Medical term                  & Language used to describe components and processes of the human body                                                                                  & what is the scientific name for the shoulder bone            & shoulder blade                & scapula                            \\\midrule
Multiple correct answers      & There is no single definite answer                                                                                                                    & city belonging to mid west of united states                  & Des Moines                    & kansas city                        \\\midrule
Spatial                       & Requires an understanding of the concept of space, location, or   proximity                                                                           & where was the tv series pie in the sky filmed                & Marlow in Buckinghamshire     & bray studios                       \\\midrule
Synonyms                      & Gold label and prediction are synonymous                                                                                                              & what is the purpose of a chip in a debit card                & control access to a resource  & security                           \\\midrule
Biological term               & Of or relating to biology or life and living processes                                                                                             & where is the ground tissue located in plants                 & in regions of new growth      & cortex                             \\\midrule
Wrong gold label              & The ground-truth label is incorrect                                                                                                                   & how do you call a person who cannot speak                    & sign language                 & mute                               \\\midrule
Wrong label                   & The human judgement is incorrect                                                                                                                      & who wrote the words to the original pledge of allegiance     & Captain George Thatcher Balch & Francis Julius Bellamy             \\\midrule
Incomplete answer             & The gold label answer contains only a subset of the full answer                                                                                       & what are your rights in the first amendment                  & religion                      & freedom of the press               \\ \bottomrule
\end{tabular}
\caption{Category definitions and examples from annotated NQ-open dataset.}
\label{categories}
\end{table*}

\section{Implementation details}
The original bi-encoder applied in \cite{risch2021semantic} uses the multi-lingual \href{https://huggingface.co/T-Systems-onsite/cross-en-de-roberta-sentence-transformer}{T-Systems-onsite/cross-en-de-roberta-sentence-transformer} \cite{may} that is based on \href{https://huggingface.co/xlm-roberta-base}{xlm-roberta-base} which was further trained on an unreleased multi-lingual paraphrase dataset resulting in the model \href{https://huggingface.co/sentence-transformers/paraphrase-xlm-r-multilingual-v1}{paraphrase-xlm-r-multilingual-v1}. The latter was fine-tuned on an English-language STS benchmark dataset \cite{cer-etal-2017-semeval} and a machine-translated German \href{https://github.com/t-systems-on-site-services-gmbh/german-STSbenchmark}{STS benchmark}.
Similar to the bi-encoder approach, the English SAS cross-encoder model relies on 
\href{https://huggingface.co/cross-encoder/stsb-roberta-large}{cross-encoder/stsb-roberta-large} which was trained on the same English STS benchmark. For German,
a new cross-encoder model had to be trained, as there were no German cross-encoder models available. It is based on deepset's \href{https://huggingface.co/deepset/gbert-large}{gbert-large} \cite{chan-etal-2020-germans} and trained on the same machine-translated German STS benchmark as the bi-encoder model, resulting in \href{https://huggingface.co/deepset/gbert-large-sts}{gbert-large-sts}.

\subsection{Hyperparameter tuning}
We did an automatic hyperparameter search \autoref{hyperparameters} for 5 trials with Optuna \cite{akiba2019optuna}. Note that cross-validation is an approximation of Bayesian optimization, so it is not necessary to use it with Optuna. The following set of hyperparameters was found to be the best: {'batch': 64, 'epochs': 2, 'warm': 0.45}.

\begin{figure*}
\centering 
\begin{minipage}[t]{0.33\textwidth}
  \includegraphics[width=\linewidth]{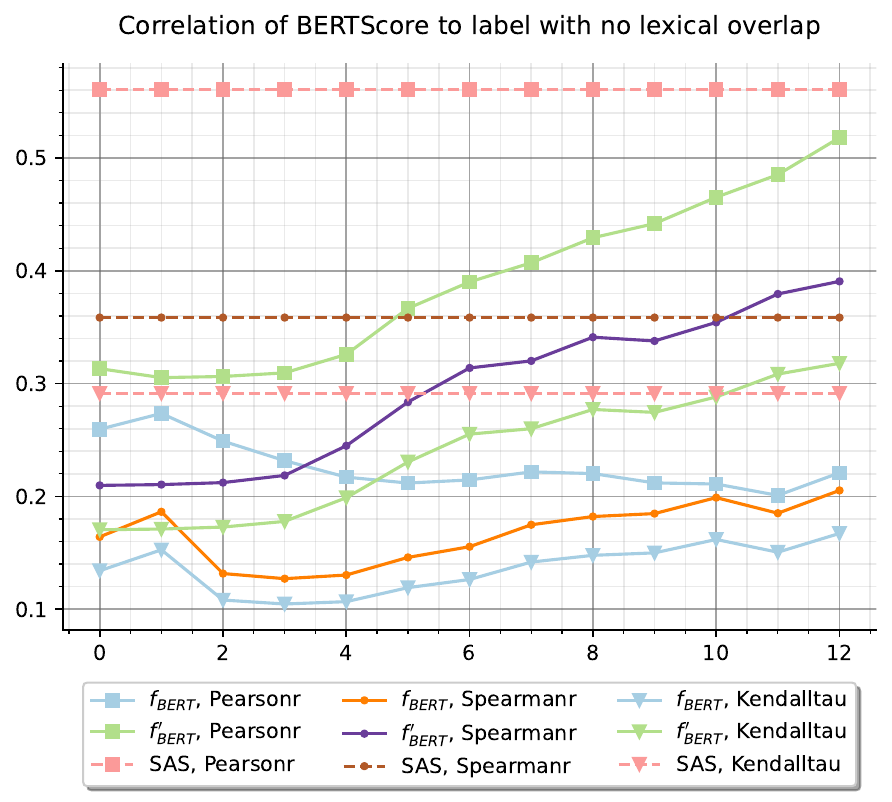}
  \label{fig:first}
\end{minipage}%
\hfill 
\begin{minipage}[t]{0.33\textwidth}
\includegraphics[width=\linewidth]{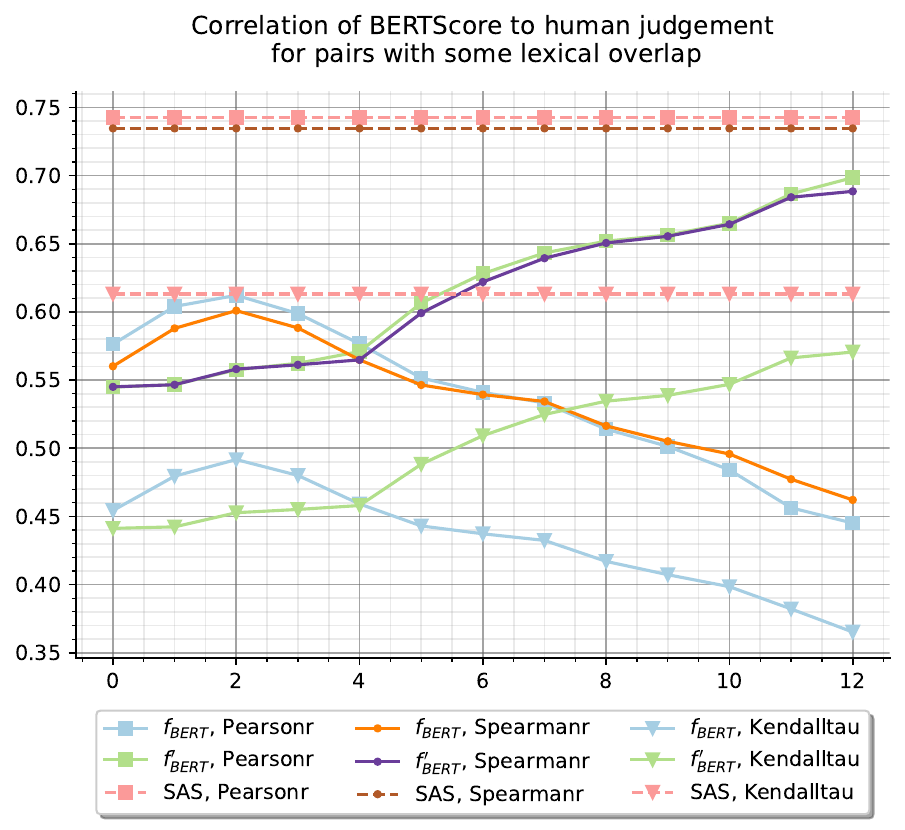}
  \label{fig:second}
\end{minipage}%
\hfill
\begin{minipage}[t]{0.33\textwidth}
  \includegraphics[width=\linewidth]{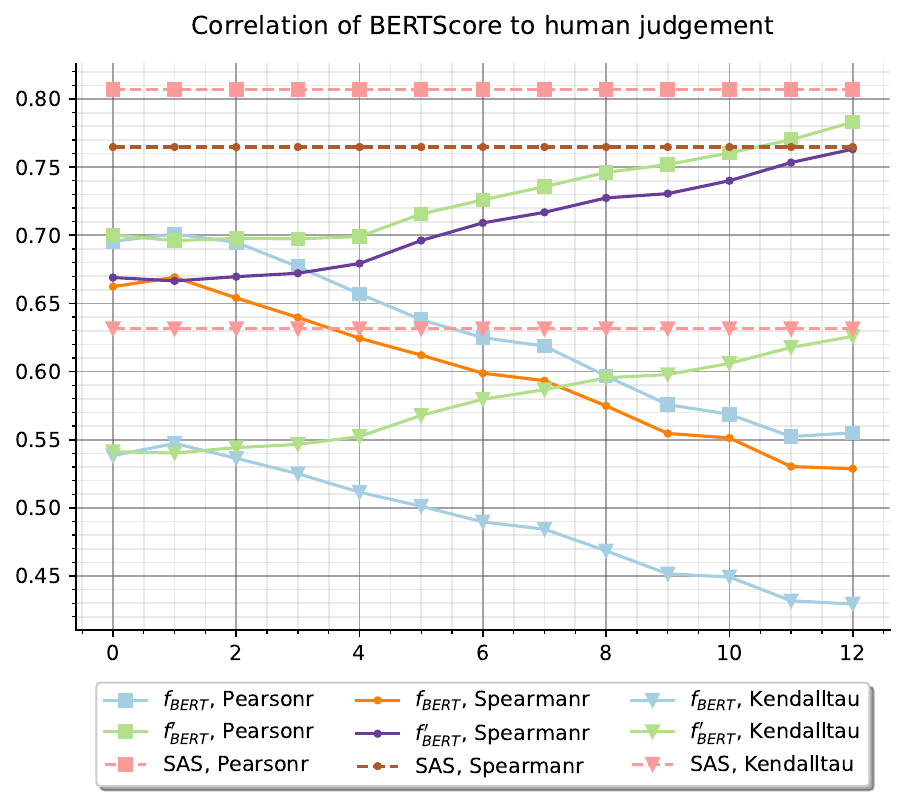}
  \label{fig:third}
\end{minipage}%
\caption{Pearson, Spearman's, and Kendall's rank correlations for different embedding extractions for when there is no lexical overlap ($F_1 = 0$), when there is some overlap ($F_1 \neq 0$) and aggregated for the SQuAD subset. $f_{BERT}$ is BERTScore vanilla and $f^{\prime}_{BERT}$ is BERTScore trained.}
\label{embedding_extraction}
\end{figure*}

To be able to use BERTScore \cite{zhang2019bertscore}, we made minor changes to accommodate for missing key-value pairs for the \cite{may} model type. 
In \autoref{embedding_extraction}, we analyse SQuAD subset dataset of answers and we observe a similar phenomenon as in \cite{risch2021semantic} when there is no lexical overlap between the answer pairs: the higher in layers we go in case of BERTScore trained, the higher the correlation values with human labels are. Quite the opposite is observed in the case of BERTScore vanilla, where it is either not as sensitive to embedding representations in case of no lexical overlap or correlations decrease with higher embedding layers. 
We did an automatic hyperparameter search for 5 trials with Optuna \cite{akiba2019optuna}. Note that cross-validation is an approximation of Bayesian optimization, so it is not necessary to use it with Optuna. We found the following best hyperparameters: {'Batch': 64, 'Epochs': 2, 'warm': 0.45}. 
\begin{table}[ht]
\begin{tabular}{@{}ll@{}}
\toprule
Batch Size & \{16, 32, 64, 128, 256\} \\ 
Epochs     & \{1, 2, 3, 4\}           \\
warm       & uniform(0.0, 0.5)        \\ \bottomrule
\end{tabular}
\caption{ Experimental setup for hyperparameter tuning of cross-encoder augmented BERTScore.}
\label{hyperparameters}
\end{table}

\begin{table*}[ht]
\resizebox{\textwidth}{!}{
\begin{tabular}{@{}lllllll@{}}
\toprule
                                & \textbf{\thead{deepset/\\gbert-large-sts}  }           & \textbf{\thead{cross-encoder/\\stsb-roberta-large } }     
                                & \textbf{\thead{T-Systems-onsite/\\cross-en-de-roberta\\-sentence-transformer}} & \textbf{bert-base-uncased}     & \textbf{\thead{deepset/\\gelectra-base} }      & \textbf{\thead{Augmented \\cross-en-de-roberta\\-sentence-transformer}}\\ \midrule
hidden\_size                    & 1,024                               & 1,024                                  & 768                                                       & 768                   & 768                        &768 \\
intermediate\_size              & 4,096                               & 4,096                                  & 3,072                                                     & 3,072                 & 3,072                      & 3,072 \\
max\_position\_embeddings       & 512                                 & 514                                    & 514                                                       & 512                   & 512                        & 514 \\
model\_type                     & bert                                & roberta                                & xlm-roberta                                               & bert                  & electra                     &xlm-roberta\\
num\_attention\_heads           & 16                                  & 16                                     & 12                                                        & 12                    & 12                          &12\\
num\_hidden\_layers             & 24                                  & 24                                     & 12                                                        & 12                    & 12                          &12\\
vocab\_size                     & 31,102                              & 50,265                                 & 250,002                                                   & 30,522                & 31,102                      &250,002\\

transformers\_version	&4.9.2	&-	&-	&4.6.0.dev0	&-	&4.12.2\\
\bottomrule
\end{tabular}}
\caption{Configuration details of each of the models used in evaluations. The architectures for the first two models and our own model follow corresponding sequence classification. T-systems-onsite model as well as our trained model follow \texttt{XLMRobertaModel}, and the other two - \texttt{BertForMaskedLM} \& \texttt{ElectraForPreTraining} archictectures respectively. Most of the models use absolute position embedding.}
\label{configs}
\end{table*}

\section{Numeric errors}
\label{numeric_errors}
Presumably, numbers are difficult to evaluate (for all metrics), including for the underlying QA model of the predictions because we observe a high amount of label 0 cases where the prediction needed to be a number, however the labels in NQ-open are not entirely reliable, more so when they are 0. Therefore, we performed two experiments using NQ-open dataset where we remove all numbers from both ground-truth answers and predictions, and in the second experiment we remove numbers only from ground-truth answers. 
We further investigated whether numbers and digit are bringing the SAS performance down. We derived a new dataset from NQ-open where any row with a number in ground-truth is removed and then evaluated the four metrics. The removal of numbers further deteriorated the SAS performance, as evident in \autoref{nq_perf_with_digit}. 

A similar experiment with SQuAD dataset shows a similar behaviour that SAS performed poorly compared to BERT-trained and Bi-Encoder metrics, but we did not observe a significant drop in performance when rows with numbers in ground-truth are removed from SQuAD since numbers are found only in 13\% of SQuAD data compared to 28\% of NQ-Open data. 

\begin{table}[ht]
\centering
\footnotesize
\setlength{\tabcolsep}{1.1\tabcolsep}
\begin{tabular}{@{}lllllll@{}}
\toprule
& \multicolumn{6}{c}{\textbf{NQ-open}}\\
& \multicolumn{3}{c}{\textbf{$F_1 = 0$}} & \multicolumn{3}{c}{\textbf{$F_1 \neq 0$}} \\
 \cmidrule(lr){2-3} \cmidrule(lr){4-5}
 \textbf{Metrics}       & $w\_num$      & $wo\_num$     &  $w\_num$    & $wo\_num$& \\ \midrule
$f_{BERT}$              & 10.9       & 13.5         & 7.1       & 22.6  \\
Bi-Encoder              & 13.3        & 13.1        & 29.9      & 25.8    \\
$f^{\prime}_{BERT}$     & 14.4        & 14.0        & 29.8      & 25.9    \\
SAS                     & 11.3        & 9.7        & 41.3      & 35.1    \\ \bottomrule
\end{tabular}
\caption{Kendall's performance on NQ-open dataset, with and without numbers.}
\label{nq_perf_with_digit}
\end{table}

\begin{table}[ht]
\centering
\footnotesize
\setlength{\tabcolsep}{1.1\tabcolsep}
\begin{tabular}{@{}lllllll@{}}
\toprule
& \multicolumn{6}{c}{\textbf{SQuAD}}\\
& \multicolumn{3}{c}{\textbf{$F_1 = 0$}} & \multicolumn{3}{c}{\textbf{$F_1 \neq 0$}} \\
 \cmidrule(lr){2-3} \cmidrule(lr){4-5}
 \textbf{Metrics}       & $w\_num$      & $wo\_num$     &  $w\_num$    & $wo\_num$& \\ \midrule
$f_{BERT}$              & 8.5       & 8.3         & 46.9       & 49.9  \\
Bi-Encoder              & 29.2        & 31.4        & 56.0      & 56.8    \\
$f^{\prime}_{BERT}$     & 30.5        & 32.7        & 56.3      & 56.7    \\
SAS                     & 27.6        & 28.4        & 60.5      & 60.8    \\ \bottomrule
\end{tabular}
\caption{Kendall's performance on SQuAD dataset, with and without numbers.}
\label{squad_perf_with_digit}
\end{table}
\begin{figure}[ht]
\centering
\small
\fbox{
\begin{minipage}{23em}
\textbf{Question}: Who killed Natalie and Ann in Sharp Objects? \\
\textbf{Ground-truth answer}: Amma\\
\textbf{Predicted Answer}: Luke  \\
\textbf{EM}: $0.00$  \\
$\mathbf{F_1}$: $0.00$  \\
\textbf{Top-1-Accuracy}: $0.00$\\  
\textbf{SAS}: $ 0.0096$  \\
\textbf{Human Judgment}: $0$  \\
$\mathbf{f_{BERT}}$: $0.226$\\	
$\mathbf{f^{\prime}_{BERT}}$: $0.145$\\	
\textbf{Bi-Encoder}: 0.208\\
$\mathbf{\tilde{f}_{BERT}}$: $0.00$\\
\textbf{Bi-Encoder (new model)}: $-0.034$
\end{minipage}}
 \caption{Representative example from NQ-open of a question and all semantic answer similarity measurement results.}

 \label{fig:ex_answer_improv}
\end{figure}

To investigate further, we created a new numbers dataset consisting of numbers as strings and their respective digit representation (digit/string and string/digit pairs) which were manually labelled as 1. These pairs were complemented by pairs of digits and their consecutive and preceding numbers, labelled as 0.
Training the bi-encoder model on this dataset resulted in no change or worse performance, the cross-encoder model on the manually annotated dataset let to non-significant improvements. Training the bi-encoder model on the dataset with a cross-encoder derived labels led to slightly less poor performance. 

\section{Distribution of scores}
\begin{figure*}[ht]
    \centering 
\includegraphics[width=.95\textwidth]{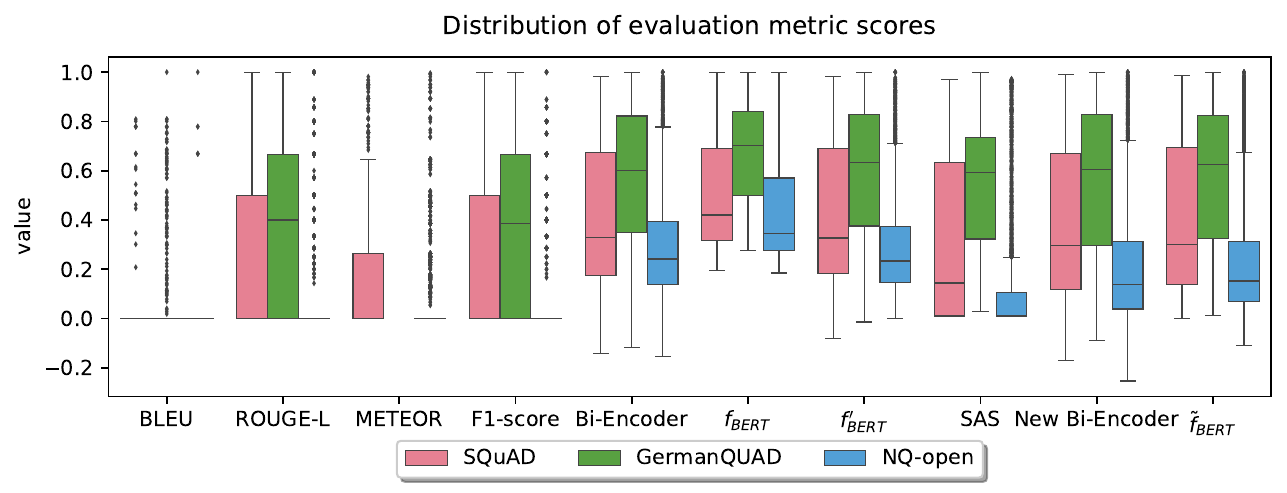}
\caption{Comparison of all (similarity) scores for the pairs in evaluation datasets. METEOR computations for GermanQuAD are omitted since it is not available for German.}

    \label{fig:comparison}
\end{figure*}

\begin{figure}[ht]
    \centering
\includegraphics[width=.45\textwidth]{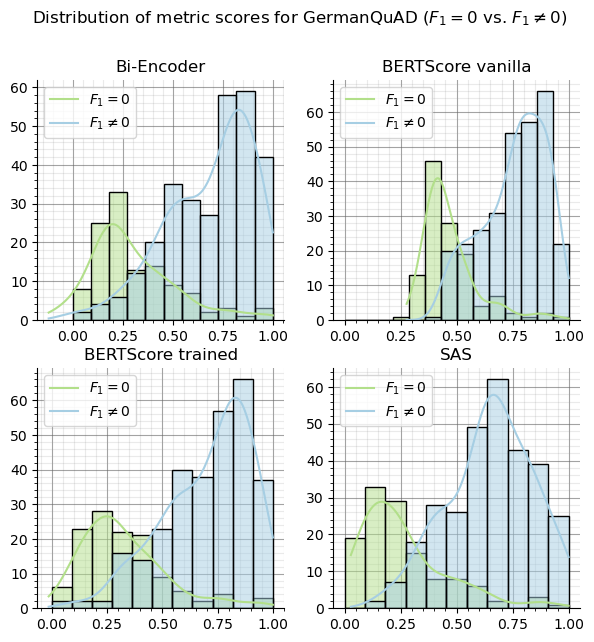}
    \caption{Distribution of scores across labels for answer-pairs in GermanQuAD.}
    \label{fig:gquad_f1_0}
\end{figure}

Score distribution for SAS and BERTScore trained shows that SAS scores are heavily tilted towards 0 (see \autoref{fig:comparison} and \autoref{fig:gquad_f1_0}).
\begin{figure}[ht]
\centering

\begin{minipage}[t]{.4\textwidth}
  \includegraphics[width=\linewidth]{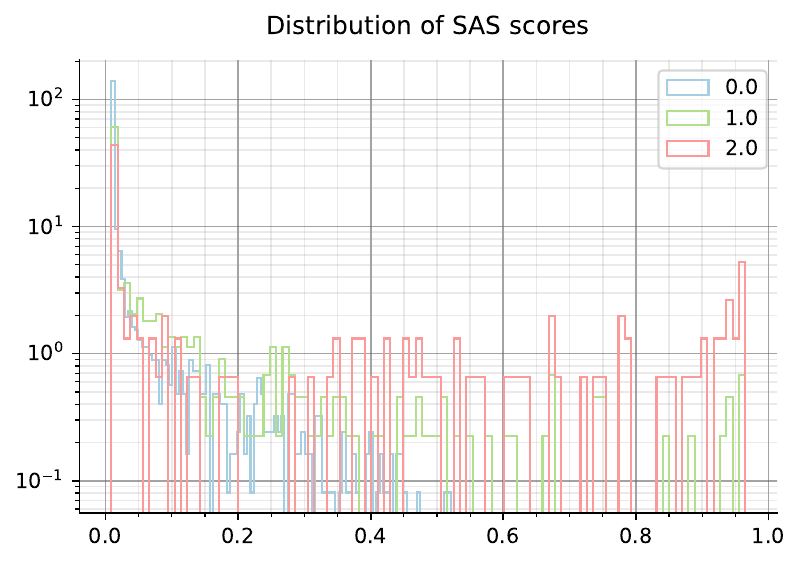}
 \end{minipage}
 \begin{minipage}[t]{0.4\textwidth}
  \includegraphics[width=\linewidth]{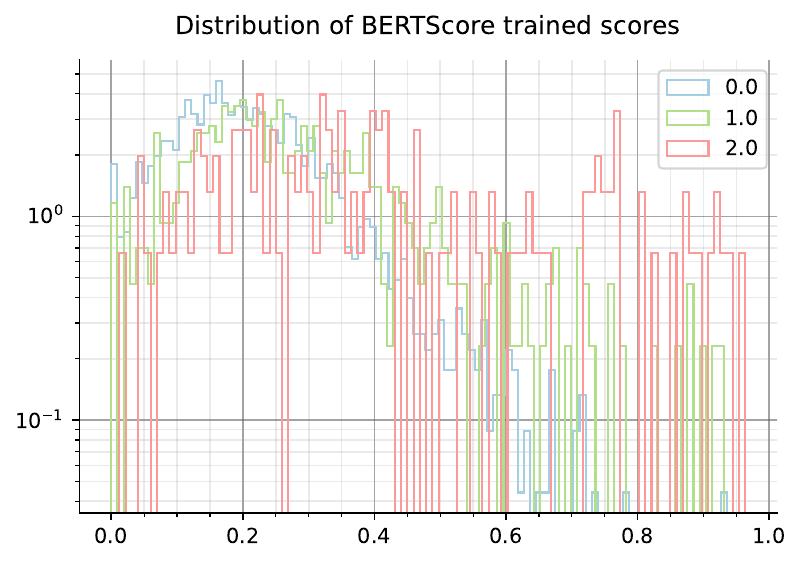}
\end{minipage}%

\label{fig:nq_open_sas_berttrained}
\caption{Distribution of SAS and BERT Trained scores for NQ-Open when $F_1 = 0$.}

\end{figure}




\section{Model Complexity}
\label{complexity}
We have scanned all metrics from \autoref{squad_and_nq_open} for time complexity on NQ-open as it is the largest evaluation dataset. Note that we haven't profiled training times as those are not defined for lexical-based metrics, but only measured CPU  time for predicting answer pairs in NQ-open. N-gram based metrics are much faster as they don't have any encoding or decoding steps involved, and they take $\thicksim$10s to generate similarity scores. The slowest is the cross-encoder as it requires concatenating answers first, followed by encoding, and it takes $\thicksim$10 minutes. Concatenation grows on a quadratic scale with the input length due to self-attention mechanism. For the same dataset, bi-encoder takes $\thicksim$2 minutes. BERTScore trained takes $\thicksim$3 minutes, hence computational costs of BERTScore and bi-encoders are comparable. 
Additional complexity for all methods mentioned above except for SAS would be marginal when used during training on the validation set. Please note the following system description: 
\newline

\fbox{
\begin{minipage}{18em}
\textbf{System} ='Darwin',  \textbf{Release}='20.6.0', \\
\textbf{Machine}='x86\_64', \textbf{Total Memory}=8.00GB,\\ 
\textbf{Total cores}=4, \textbf{Frequency}=2700.00Mhz\\
\end{minipage}}

\end{document}